\documentclass[11pt,a4paper]{article}
\usepackage[hyperref]{emnlp2020}
\usepackage{times}
\usepackage{latexsym}

\usepackage{microtype}

\aclfinalcopy 

\usepackage[utf8]{inputenc}
\usepackage{booktabs}
\usepackage{tabularx}

\usepackage{array}
\usepackage{graphicx}
\usepackage{todonotes}
\usepackage{xcolor}
\usepackage{subfig}
\usepackage{multirow}
\usepackage{amsfonts}
\usepackage{bbm}
\usepackage{enumitem}
\usepackage{fdsymbol}

\newcolumntype{L}[1]{>{\raggedright\arraybackslash}p{#1}}

\definecolor{darkgreen}{rgb}{0.09, 0.45, 0.27}
\definecolor{darkviolet}{rgb}{0.58, 0.0, 0.83}

\usepackage{color}

\title{Uncertainty over Uncertainty: Investigating the Assumptions, Annotations, and Text Measurements of Economic Policy Uncertainty}

\author{Katherine A.~Keith\thanks{~~This work was done during an internship at Bloomberg.} \\
  University of Massachusetts Amherst \\
  \texttt{kkeith@@cs.umass.edu} \\
  \And
  Christoph Teichmann \\
  Bloomberg \\
  \texttt{cteichmann1@bloomberg.net} \\
  \AND
  Brendan O'Connor \\
  University of Massachusetts Amherst \\
  \texttt{brenocon@@cs.umass.edu} \\
  \And
  Edgar Meij \\
  Bloomberg\\
  \texttt{emeij@bloomberg.net}\\
  }

\date{}

\begin{document}
\maketitle

\begin{abstract}
Methods and applications are inextricably linked in science, and in particular in the domain of text-as-data. 
In this paper, we examine one such text-as-data application, an established economic index that measures \emph{economic policy uncertainty} from keyword occurrences in news.
This index, which is shown to correlate with firm investment, employment, and excess market returns, has had substantive impact in both the private sector and academia. 
Yet, as we revisit and extend the original authors' annotations and text measurements we find interesting text-as-data methodological research questions: 
(1) Are annotator disagreements a reflection of ambiguity in language?
(2) Do alternative text measurements correlate with one another and with measures of external predictive validity? 
We find for this application (1) some annotator disagreements of \emph{economic policy uncertainty} can be attributed to ambiguity in language, and (2) switching measurements from keyword-matching to supervised machine learning classifiers results in low correlation, a concerning implication for the validity of the index.   
\end{abstract}


\section{Introduction}

The relatively novel research domain of \emph{text-as-data}, which uses computational methods to automatically analyze large collections of text, is a rapidly growing subfield of computational social science with applications in political science~\cite{grimmer2013text}, sociology~\cite{evans2016machine}, and economics~\cite{gentzkow2019text}. 
In economics, textual data such as news editorials \cite{tetlock2007giving}, central bank communications~\cite{lucca2009measuring}, financial earnings calls~\cite{keith2019modeling}, company disclosures~\cite{hoberg2016text}, and newspapers~\cite{thorsrud2020words} have recently been used as new, alternative data sources.

{\renewcommand\normalsize{\small}%
\normalsize
\begin{table*}[t]
  \centering
      \begin{tabularx}{\linewidth}{lX}
      \toprule 
      No. & Example  \\ 
      \midrule
 		1 & 
 		\vspace{-0.3cm}
 		{\color{darkviolet} Demand for new clothing} is {\color{red} uncertain} {\color{orange} because} {\color{blue} several states may implement large hikes in their sales tax rates}.  \\ \hline 

 		2 & The outlook for the {\color{blue} H1B visa program} {\color{red} remains highly uncertain}. {\color{orange} As a result}, some high-tech firms fear that {\color{darkviolet} shortages of qualified workers will cramp their expansion plans}.  \\ \hline

 		3 & 
 		\vspace{-0.2cm} 
 		{\color{blue} The looming political fight over whether to extend the Bush-era tax cuts} {\color{orange} makes} it {\color{red} extremely difficult to forecast} {\color{darkviolet} federal income tax collections in 2011}.\\ \hline

 		4 & 
 		\vspace{-0.2cm}
 		{\color{red} Uncertainty} about {\color{blue} prospects for war in Iraq} {\color{orange} has encouraged} a build-up of petroleum inventories and {\color{darkviolet} pushed oil prices higher}.\\ \hline


 		5 & Some economists claim that {\color{red} uncertainties} due to {\color{blue} government industrial policy in the 1930s} {\color{orange} prolonged and deepened} {\color{darkviolet} the Great Depression}. \\ \hline

 		6 & 
 		\vspace{-0.2cm}
 		{\color{red} It remains unclear} whether the {\color{blue} government will implement new incentives} for {\color{darkviolet}small business hiring}. \\
      \bottomrule
  \end{tabularx}
\caption{
      Positive examples of \emph{policy-related economic uncertainty.} 
      We label spans of text as indicating {\color{blue} policy}, {\color{darkviolet} economy}, {\color{red} uncertainty}, or a {\color{orange} causal relationship}.
      Examples were selected from hand-labeled positive examples and the coding guide provided by \citet{baker2016measuring}.  \label{t:exs}
    }
\end{table*}}

In one such economic text-as-data application, \citet{baker2016measuring} aim to construct an \emph{economic policy uncertainty} (EPU) index whereby they quantify the aggregate level that policy is influencing economic uncertainty (see Table \ref{t:exs} for examples).
They operationalize this as the proportion of newspaper articles that match keywords related to the economy, policy, and uncertainty. 

The index has had impact both on the private sector and academia.\footnote{As of October 7, 2020, Google Scholar reports \citet{baker2016measuring} to have over 4400 citations.}
In the private sector, financial companies such as Bloomberg, Haver, FRED, and Reuters carry the index and sell financial professionals access to it.
Academics show economic policy uncertainty has strong relationships with other economic indicators:
\citet{gulen2016policy} find a negative relationship between the index and firm-level capital investment, and 
\citet{brogaard2015asset} find that the index can positively forecast excess market returns.

The EPU index of \citeauthor{baker2016measuring} has substantive impact and is a real-world demonstration of finding economic signal in textual data. 
Yet, as the subfield of text-as-data grows, so too does the need for rigorous methodological analysis of how well the chosen natural language processing methods operationalize the social science construct at hand.
Thus, in this paper we seek to re-examine \citeauthor{baker2016measuring}'s linguistic, annotation, and measurement assumptions. 
Regarding measurement, although keyword look-ups yield high-precision results and are interpretable, they can also be brittle and may suffer from low recall. \citeauthor{baker2016measuring} did not explore alternative text measurements based on, for example, word embeddings or supervised machine learning classifiers. 

In exploring \citeauthor{baker2016measuring}'s construction of EPU, we identify and disentangle multiple sources of uncertainty. 
First, there is the \emph{real underlying uncertainty} about economic outcomes due to government policy that the index attempts to measure. Second, there is \emph{semantic uncertainty} that can be expressed in the language of newspaper articles. Third, there is \emph{annotator uncertainty} about whether a document should be labeled as EPU or not. Finally, there is \emph{modeling uncertainty} in which text classifiers are uncertain about the decision boundary between positive and negative classes. 

In this paper, we revisit and extend \citeauthor{baker2016measuring}'s human annotation process (\S\ref{s:annotator}) and computational pipeline that obtains EPU measurement from text (\S\ref{s:mesurment}).
In doing so, we draw on concepts from quantitative social science's \emph{measurement modeling}, mapping observable data to theoretical constructs, which emphasizes the importance of \emph{validity} (is it right?) and \emph{reliability} (can it be repeated?) \cite{loevinger1957objective,messick1987validity,quinn2010analyze,jacobs2019measurement}. 

Overall, this paper contributes the following:
\begin{itemize}[leftmargin=*]
    \item We examine the assumptions  \citeauthor{baker2016measuring} use to operationalize \emph{economic policy uncertainty} via keyword-matching of newspaper articles. We demonstrate that using keywords collapses some rich linguistic phenomena such as \emph{semantic uncertainty} (\S\ref{ss:semantic-uncertainty}).
    \item We also examine the \emph{causal} assumptions of \citeauthor{baker2016measuring} through the lens of \emph{structural causal models} \cite{pearl2009causality} and argue that readers' \emph{perceptions} of economic policy uncertainty may be important to capture (\S\ref{ss:causal}).
    \item We conduct an annotation experiment by re-annotating documents from  \citeauthor{baker2016measuring}. We find preliminary evidence that disagreements in annotation could be attributed to inherent ambiguity in the language that expresses EPU (\S\ref{s:annotator}). 
    \item Finally, we replicate and extend \citeauthor{baker2016measuring}'s data pipeline with numerous \emph{measurement sensitivity} extensions: filtering to US-only news, keyword-matching versus supervised document classifiers, and prevalence estimation approaches. 
    We demonstrate that a measure of \emph{external predictive validity}, i.e.,  correlations with a stock-market volatility index (\emph{VIX}), is particularly sensitive to these decisions (\S\ref{s:mesurment}).
\end{itemize}

{\renewcommand\normalsize{\small}%
\normalsize
{\tiny
\begin{table*}[t]
  \centering
      \begin{tabular}{l >{\raggedright\arraybackslash}p{6.5cm}
      >{\raggedright\arraybackslash}p{6.5cm}}
      \toprule
               & KeyOrg  & KeyExp \\
      \toprule
      Economy     & economic, economy  & + growth, economies, financial, recession, slowdown  \\
      \hline
      Uncertainty & uncertain,
uncertainty
 &  + unclear, unsure, uncertainties, turmoil, confusion, worries
 \\
      \hline
      Policy &
      \multicolumn{2}{L{13cm}}
      {regulation, deficit, legislation, congress, white house, federal reserve, the fed, regulations, regulatory, deficits, congressional, legislative, legislature} 
 \\
      \bottomrule
  \end{tabular}
  \caption{Original keywords used in \citeauthor{baker2016measuring}'s monthly United States index (KeyOrg). Expanded keywords include all words from KeyOrg plus the five nearest neighbors from pre-trained GloVe embeddings for the economy and uncertainty categories (KeyExp). \label{t:epu-keywords}}
\end{table*}
}}

\section{Assumptions of Measuring \emph{Economic Policy Uncertainty} from News}

The goal of \citet{baker2016measuring} is to measure the \emph{theoretical construct} of \emph{policy-related economic uncertainty} (EPU) for particular times and geographic regions. \citeauthor{baker2016measuring} assume they can use information from newspaper articles as a \emph{proxy} for EPU, an assumption we explore in great detail in Section~\ref{ss:causal}, and they define EPU very broadly in their coding guidelines: ``Is the article about policy-related aspects of economic uncertainty, even if only to a limited extent?"\footnote{\url{http://policyuncertainty.com/media/Coding_Guide.pdf}} For an article to be annotated as positive, there must be a stated causal link between \emph{policy} and \emph{economic consequences} and either the former or the latter must be \emph{uncertain}.\footnote{``If the article discusses economic uncertainty in one part and policy in another part but never discusses policy in connection to economic uncertainty, then do not code it as about economic policy uncertainty."}
 Grounds for labeling a document as a positive include ``uncertainty regarding the economic effects of policy actions" (or inactions), and ``uncertainty over who makes or will make policy decisions that have economic consequences." In Table~\ref{t:exs}, we provide examples of text spans that successfully encode EPU given these guidelines. For instance, the first example indicates that a government policy (increase in state sales tax) is causing uncertainty in the economy (demand for new clothing).
 \citeauthor{baker2016measuring} \emph{operationalize} this theoretical construct of EPU as  keyword-matching of newspaper documents: for each document, if the document has at least one word in each of the economy, uncertainty, and policy keyword categories (see Table~\ref{t:epu-keywords} in the Appendix) then it is considered a positive document. Counts of positive documents are summed and then normalized by the total number of documents published by each news outlet.



\subsection{Semantic Uncertainty}\label{ss:semantic-uncertainty}
While the keywords
\citet{baker2016measuring} select (``{uncertain}'' or ``{uncertainty}'') are the most overt ways to express uncertainty via language, they do not capture the full extent of how humans express uncertainty. For instance, Example No.~6 in Table~\ref{t:exs} would be counted as a negative by \citeauthor{baker2016measuring} despite indicating semantic uncertainty via the phrase ``it remains unclear." 
These keyword assumptions are a threat to \emph{content validity}, ``the extent to which a measurement model captures everything we might want it to" \cite{jacobs2019measurement}. 

We look to definitions from linguistics to potentially expand the operationalization of uncertainty; we refer the reader to \citet{szarvas2012cross} for all subsequent definitions and quotes.
In particular,
\emph{uncertainty} is defined as a phenomenon that represents a lack of information. 
With respect to truth-conditional semantics, \emph{semantic uncertainty} refers to propositions ``for which no truth value can be attributed given the speaker's mental state.''  
 \emph{Discourse-level uncertainty} indicates ``the speaker intentionally omits some information from the statement, making it vague, ambiguous, or misleading'' and in the context of \citeauthor{baker2016measuring}
could result from journalists' linguistic choices to express ambiguity in economic policy uncertainty. For instance, in the first example in Table~\ref{t:sem-exs}, the lexical cues ``suggest" and ``might" indicate to the reader that the journalist writing the article is unclear about the intention of Alan Greenspan. In contrast, \emph{epistemic modality} ``encodes how much certainty or evidence a speaker has for the proposition expressed by his utterance,'' (e.g., ``Congresswoman X: `We \emph{may} delay passing the tariff bill.'") and  
\emph{doxastic modality} refers to the beliefs of the speaker (``I \emph{believe} that Congress will \dots"). In the second example in Table~\ref{t:sem-exs}, the entity ``he" seems to be uncertain about the fate of the economy because he ``shakes his head in bewilderment," which demonstrates that uncertainty can also be conveyed through world knowledge and inference. 

Collapsing all these types of \emph{semantic uncertainty} to the keywords ``uncertainty" and ``uncertain" has major implications: (a) the relationship between the uncertainty journalists express and what readers infer impacts the causal assumptions (\S\ref{ss:causal}) and annotation decisions (\S\ref{s:annotator}) of this task, and (b) \citeauthor{baker2016measuring}'s keywords are most likely low-recall which could affect empirical measurement results (\S\ref{s:mesurment}). We see fruitful future work in improving \emph{content validity} and recall via automatic uncertainty and modality analysis from natural language processing, e.g.~\citet{mcshane2004mood,ganter2009finding,sauri2009factbank,farkas2010conll,szarvas2012cross}.

{\renewcommand\normalsize{\small}%
\normalsize
\begin{table}[t]
  \centering
      \begin{tabularx}{\columnwidth}{Xl}
      \toprule
      Example & Docid \\
      \midrule
    The stock market had soared on Mr. Greenspan's suggestion that global financial problems posed as great a threat to the United States as inflation did, \textbf{suggesting} that a rate cut to stimulate the economy \textbf{might be on the horizon} & 1047100 \\ 
    \midrule 
      But ask him whether the Mexican stock market will rise or plunge tomorrow and \textbf{he shakes his head in bewilderment.} & 1043578 \\
      \bottomrule 
  \end{tabularx}
  \caption{
        Selected examples extracted from the New York Times Annotated Corpus (\emph{NYT-AC}) that convey semantic uncertainty about the economy. Bolding is our own. Docids are from the \emph{NYT-AC} metadata. \label{t:sem-exs}
      }
\end{table}
}
\vspace{-0.2cm} 
\subsection{Causal Assumptions}\label{ss:causal}
\begin{figure}[t]
\centering
\includegraphics[width=0.7\columnwidth]{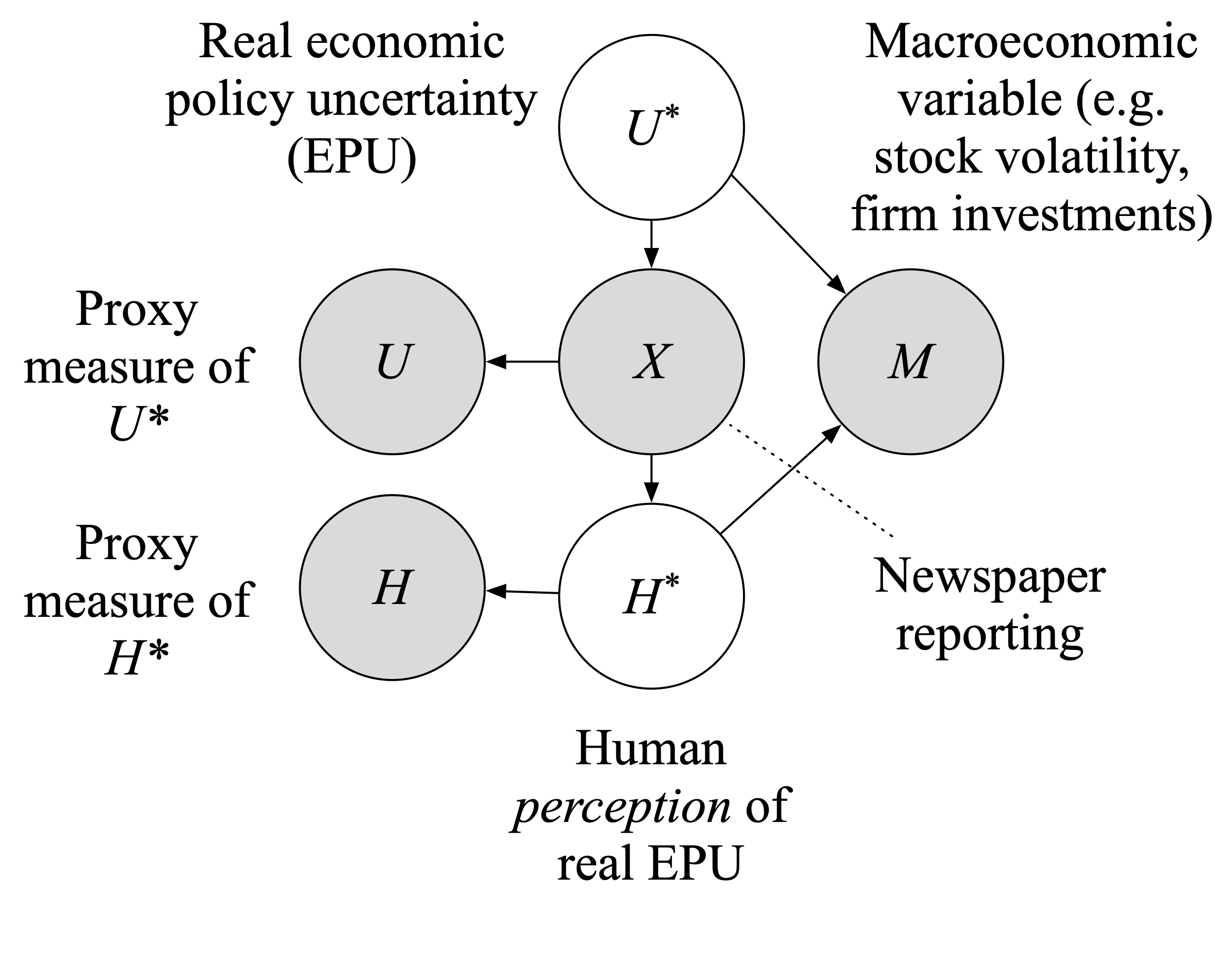}
\caption{Structural causal model of the \emph{economic policy uncertainty} measurements in which variables are nodes and directed edges denote causal dependence. Unlike \citet{baker2016measuring} who claim to measure $U$, we posit that measuring $H$ is important. Shaded nodes are observed variables and unshaded nodes are latent. \label{f:causal}}
\end{figure}

\vspace{-0.1cm} 
Using the paradigm of \emph{structural causal models} \cite{pearl2009causality}, we re-examine the causal assumptions of \citeauthor{baker2016measuring}. 
In Figure~\ref{f:causal}, for a single time-step,\footnote{\citet{baker2016measuring} aggregate by day, month, quarter, or year.} $U^*$ represents the real, aggregate level of economic policy uncertainty in the world which is unobserved. If one could obtain a measurement of $U^*$, then one could analyze the causal relationship between $U^*$ and other macroeconomic variables, $M$. 
Presumably, newspaper reporting, $X$, is affected by $U^*$ and $x = f_{X} (u^*)$ where $f_X$ is a non-parametric function that represents a causal process. In our setting, $f_X$ represents the process of media production: for example, the ability of journalists to collect information from sources; or editorial decisions on what topics will be published.
The major assumption of \citeauthor{baker2016measuring} is that they can obtain a measure of $U^*$ via a \emph{proxy} measure from newspaper text, $U$, where $u = f_U(x)$. By simple composition, $u = f_U(f_X(u^*))$. Yet, aside from examining the political bias of media, \citeauthor{baker2016measuring} largely ignore $f_X$ and how the media production process could influence EPU measurements.

However, an alternative causal path from $U^*$ to $M$ goes through $H^*$, the macro-level human perception of real EPU.
In this case, $U^*$ is irrelevant as long as people are \emph{perceiving} policy-related economic uncertainty to be changing, they could potentially make real economic decisions (e.g.~hiring or purchases) that could affect the greater macro-economy, $M$.

It is unclear how to design a causal intervention in which one manipulates the real EPU, $do(U^*)$, in order to estimate its effect on $X$ and $M$. However, one could design an ideal causal experiment to intervene on newspaper text, $do(X)$; one could artificially change the level of EPU coverage in synthetic articles, show these to participants, and measure the resulting difference in participants' economic decisions.
If $H^*$ to $M$ is the causal path of interest,\footnote{There is some evidence from the original authors that human perception is important: In the EPU index released to the public, one of three underlying components is a disagreement of economic forecasters as a proxy for uncertainty. See \url{http://policyuncertainty.com/methodology.html}.}
then it is extremely important to measure and model human  \emph{perception} of EPU, an assumption we explore in terms of annotation decisions in Section~\ref{s:annotator}. 

{\renewcommand\normalsize{\small}%
\normalsize
{\small
\begin{table*}[t]
  \centering
  
      \begin{tabular}{l 
      >{\raggedright\arraybackslash} p{1.2cm}
      >{\raggedleft\arraybackslash} p{1.2cm} 
      >{\raggedleft\arraybackslash} p{1.2cm} >{\raggedleft\arraybackslash} p{1.6cm} >{\raggedleft\arraybackslash} p{1.8cm} >{\raggedleft\arraybackslash} p{1.2cm} r}
      \toprule
      Subset & Ann. Source & Num. Docs & Num. Anns. & Prop. Pos. Anns. & Prop. Docs. Agr. & Pairwise Agree & Krip.-$\alpha$ \\
      \toprule
      All & BBD &13797&16060&0.42&--&--&--\\ 
2+ Anns. & BBD &2150&4413&0.43&0.80&0.80&0.60\\  \midrule
Sample A (Unan.) & BBD & 19&41&0.29&1.00&1.00&1.00\\ 
& Ours &19&96&0.29&0.37&0.68&0.21\\ 
Sample B (Non-unan.) & BBD &18&40&0.50&0.00&0.07&-0.80\\ 
& Ours &18&97&0.54&0.28&0.65&0.27\\
\bottomrule
  \end{tabular}
  
  \caption{
  \textbf{Rows 1-2:} Descriptive statistics for \emph{BBD}, \citet{baker2016measuring}'s annotated dataset, and the subset of these documents that have more than two annotations each (2+ Anns.). \textbf{Rows 3-6:} Sample A with \emph{unanimous} (unan.) agreement in \emph{BBD} labels and Sample B with \emph{non-unanimous} (non-unan.) \emph{BBD} labels. 
  For these samples, we gather additional annotations. \textbf{Columns:}  Annotation (ann.) source, number of documents (num.~docs), number of annotations (num.~anns.), proportion of positive annotations (prop.~positive anns.), proportion of documents for which all annotator labels are in unanimous agreement (prop.~docs.~agr.), pairwise agreement in labels, and Krippendorff's $\alpha$ (Krip.-$\alpha$). 
  \label{t:ann-exp}}
  
\end{table*}
}

\begin{table}[t]
  \centering
      \begin{tabular}{crr}
      \toprule
      Sample        & PXA & Total pairs   \\
      \toprule
      A & 0.70 & 206\\
      B & 0.50 & 218\\ 
     \bottomrule
  \end{tabular}
  \caption{Pairwise cross-agreement (PXA) rates between \emph{BBD} and our annotations. \label{t:cross}}
\end{table}}

\section{Annotator Uncertainty} \label{s:annotator}

Reliable human annotation is essential for both building supervised classifiers and assessing the internal validity of text-as-data methods. In order to validate their EPU index,  \citeauthor{baker2016measuring} sample documents from each month, obtain binary labels on the documents from annotators, and then construct a ``human-generated" index which they report has a 0.86 correlation with their keyword-based index (aggregated quarterly). Yet, in our analysis of \citeauthor{baker2016measuring}'s annotations (denoted below as {\emph{BBD}}), we find only 16\% of  documents have more than one annotator and of these, the agreement rates are moderate: 0.80 pairwise agreement and 0.60 Krippendorff's $\alpha$ chance-adjusted agreement \cite{artstein2008inter}. See Line 2 of Table~\ref{t:ann-exp} for additional descriptive statistics of these annotations. The original authors did not address whether this disagreement is a result of annotator bias, error in annotations, or true ambiguity in the text.

In contrast to the popular paradigm that one should aim for high inner-annotator agreement rates \cite{krippendorff2018content}, recent research has shown ``disagreement between annotators provides a useful signal for phenomena such as ambiguity in the text" \cite{dumitrache2018crowdsourcing}. Additionally, recent research in natural language processing \cite{paun2018comparing,pavlick2019inherent} and computer vision \cite{sharmanska2016ambiguity} has leveraged annotator uncertainty to improve modeling. Thus, for our setting, we ask the following research question:


\textbf{RQ1:} Is there inherent ambiguity in the language that expresses \emph{economic policy uncertainty}? If so, are annotator disagreements a reflection of this ambiguity?  

The following evidence lends to our hypothesis that there \emph{is} inherent ambiguity in whether documents encode EPU: (1) the original coding guide of \citeauthor{baker2016measuring} had 17 pages of ``hard calls" that describe difficult or ambiguous documents, (2) there was a moderate amount of annotator disagreement in \emph{BBD} (Table~\ref{t:ann-exp}), (3) we qualitatively analyze examples with disagreement and reason about what makes the inferences of these documents difficult (\S\ref{ss:qualitative}, and  Tables~\ref{t::disagreement_examples} and \ref{t::agreement_examples} in the Appendix), and (4) we run an experiment in which we gather additional annotations and show that our annotations have more disagreement with documents that have non-unanimous labels in \emph{BBD}  (\S\ref{ss:our-anns}). 


\vspace{-0.3cm}
\subsection{Our annotation experiment} \label{ss:our-anns}
The ideal assessment of inherent annotator uncertainty would be to gather a large number of annotations for many documents and then analyze the posterior distribution over labels.\footnote{For instance, \citet{pavlick2019inherent} analyze disagreement in natural language inference by gathering 50 annotations per document and find the label distributions are often bi-modal, indicating 
meaningful disagreement.} We perform a similar, small-scale experiment in which we recruit 10 annotators, a mix of professional analysts and PhD students, who annotate 37 documents for a total of 193 annotations.
\footnote{We originally sampled 40 documents but after annotation had to discard some that were duplicates or had errors from HTML extraction.}
We sampled documents from the pool of \emph{BBD} documents that had more than one annotator and the \emph{BBD} labels were unanimous (Sample A) and non-unanimous (Sample B).
We re-annotated these samples in order to provide insight into the nature of these unanimous and non-unanimous labels. See Figure~\ref{f:ann-instruct} in the Appendix for our full annotation instructions. 


\textbf{Pairwise cross-agreement.}
In order to quantitatively compare two annotation rounds (ours vs.~\citeauthor{baker2016measuring}'s), we provide a new metric, \emph{pairwise cross-agreement} (PXA). Formally, for each document of interest, $d \in \mathcal{D}$, let the $\mathcal{A}_d$ and $\mathcal{B}_d$ be the set of annotations on that document from each of the two rounds respectively.
Let $P_d$ be the set of all pairs, $(a \in \mathcal{A}_d, b \in \mathcal{B}_d)$ from combining one annotation from each of the two rounds. 
Then, 
\begin{equation}
    \textrm{PXA} = \frac
    {\sum_{d \in \mathcal{D}} \sum_{(a, b) \in P_d} \mathbbm{1}(a = b) }
    {\sum_{d \in \mathcal{D}} |P_d|}
    .
\end{equation}

\textbf{Results.} The results of our experiment (Tables~\ref{t:ann-exp} and \ref{t:cross}) provide evidence supporting our hypothesis that there is inherent ambiguity in documents about EPU that contributes to annotator disagreement.
 In Table~\ref{t:cross}, PXA is higher in Sample A (0.70), in which \emph{BBD} annotators had unanimous agreement, compared to Sample B (0.50) in which \emph{BBD} annotators had non-unanimous labels. Since our annotations agreed with Sample A more, this could indicate these documents inherently have more agreement. The pairwise agreement between our annotations on Sample A and B are roughly the same (Table~\ref{t:ann-exp}) but the proportion of documents that had unanimous agreement among our five annotators per document was slightly more in Sample A versus Sample B (0.37 vs.~0.28). 
Limitations of our experiment include that our sample size is relatively small and our annotation instructions are different and significantly shorter than \citeauthor{baker2016measuring}.




\subsection{Qualitative Document Analysis} \label{ss:qualitative}
Our qualitative analysis suggests that readers' \emph{perceptions} of EPU differ meaningfully and it is difficult to measure EPU with a simple document-level binary label. 
In Tables~\ref{t::agreement_examples} and \ref{t::disagreement_examples} in the Appendix, we present documents with the highest levels of agreement from Sample A and disagreement from Sample B. Annotators are likely to disagree on the label of the document when need real world knowledge to infer whether a policy is contributing to economic uncertainty. 
For instance, in Table~\ref{t::disagreement_examples} Example~1, the reader has to infer that the author of an op-ed would only write an op-ed about a policy if it was uncertain, but the uncertainty is never explicitly stated in text. In other instances, the causal link between policy and economic uncertainty is unclear. In Table~\ref{t::disagreement_examples} Example~4, economic downturn is mentioned as well as turnover in the administration but these are never explicitly linked; yet, some annotators may have read ``questions about what lies ahead" as uncertainty that also encompasses economic uncertainty.
Although there has been a rise of common sense reasoning research in natural language processing (e.g.~ \citet{bhagavatula2020abductive,huang2019cosmos,sapsocialiqa}), we suspect current state-of-the-art NLP systems would be unable to accurately resolve the inferences stated above.
Furthermore, if there is inherent ambiguity in the language that expresses EPU, and, as we argue in Section~\ref{ss:causal}, human perception is important, then we may desire to build models that can \emph{identify} ambiguous documents and account for the uncertainty from ambiguity of language into measurement predictions, e.g.~\citet{paun2018comparing}. We leave this for future work.  





\section{Measurement} \label{s:mesurment}


{\renewcommand\normalsize{\small}%
\normalsize

\begin{table}[t]
\centering
      \begin{tabular}{llrrrr}
      \toprule
      Split &Model   & Prec. & Recall & F1 & Acc. \\
      \toprule
Train & KeyOrg&0.63&0.67&0.65&0.65 \\ 
      & LogReg-BOW&0.86&0.90&0.88&0.88 \\
\hline  
Test & KeyOrg&0.61&0.69&0.64&0.70 \\ 
      &LogReg-BOW&0.69&0.72&0.71&0.76 \\ 
      \bottomrule
  \end{tabular}
  \captionof{table}{Document-level classification statistics. Training is \emph{BBD} documents 1985-2007 (N=1844) with annotations from a single annotator and testing is all \emph{BBD} annotated documents 2007-2012 (N=687). For testing, the majority class is used and ties are randomly broken. \label{t:doc-level}}
\end{table}



\begin{figure*}[!tbp]
  \centering
  \begin{minipage}[b]{0.65\textwidth}
  \subfloat{
\centering
  \includegraphics[clip,width=0.93\textwidth]{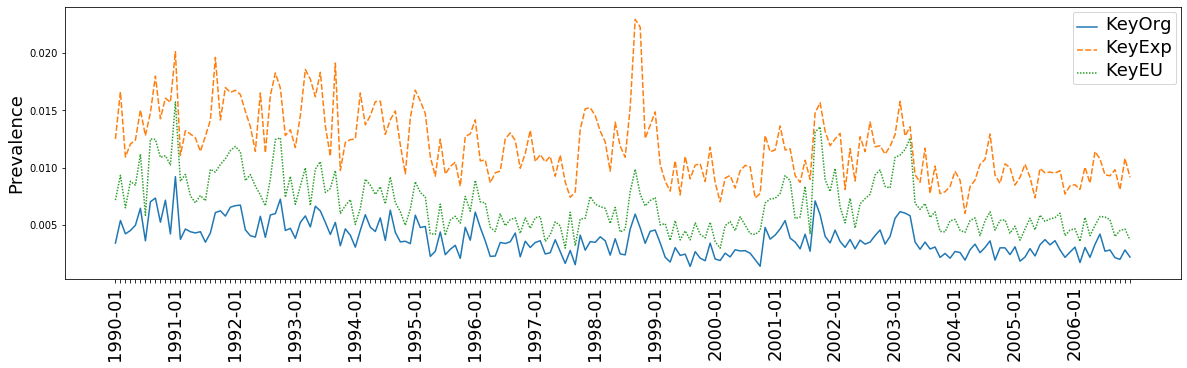}%
}
\centering

\subfloat{%
\hspace{.42cm}
  \includegraphics[clip,width=0.96\textwidth]{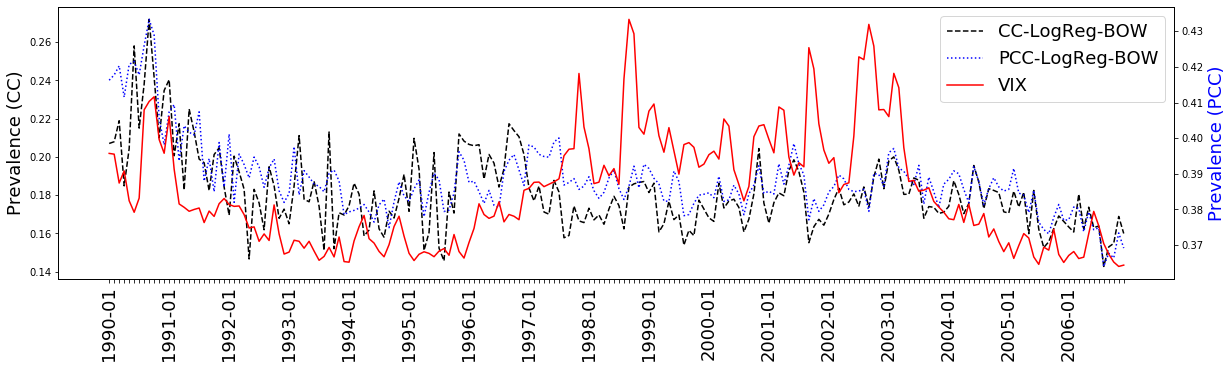}%
}
\caption{EPU Index, prevalence of documents exhibiting \emph{economic policy uncertainty}, at inference time on the \emph{NYT-AC} for all keyword methods (top) and document classifier methods (bottom) as well as the VIX. Note, for the bottom figure, the scale of the y-axis differs for CC versus PCC.  \label{f:time-series}}
  \end{minipage}
  \hfill
  \begin{minipage}[b]{0.3\textwidth}
  \centering
\includegraphics[width=1.0\textwidth]{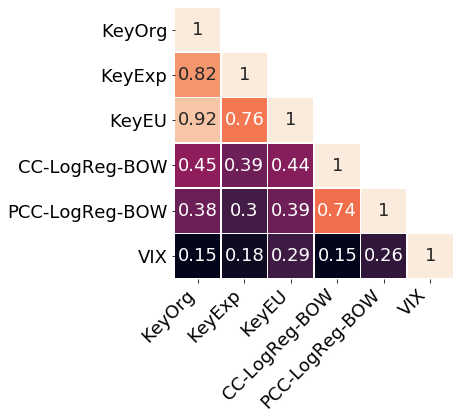}
\caption{Pearson correlation between all text measurement models and the VIX. \label{f:corr}}
  \end{minipage}
\end{figure*}}

For text-as-data applications, 
substantive results are contingent on how researchers operationalize measurement of the (latent) theoretical construct of interest via observed text data. 
Using \citeauthor{baker2016measuring}'s original causal assumptions (Section~\ref{ss:causal}), we formally define the \emph{measurement} of interest as:
\vspace{-0.3cm} 
\begin{equation}
    U = g(X),
    \vspace{-0.1cm} 
\end{equation}
where $g$ is the measurement function that maps text, $X$, to economic policy uncertainty, $U$.\footnote{\citet{egami2018make} call this $g$ function the \emph{codebook} function and describe how it can generically map text to any lower-dimensional representation.} 
For text-as-data practitioners, we emphasize that there is a ``garden of forking paths" \cite{gelman2014statistical} of how $g$ can be operationalized, for instance, in the representation of text (bag-of-words vs.~embeddings), document classification function (deterministic keyword matching vs.~supervised machine learning classifiers), and ways of aggregating individual document predictions (mean of predictions~vs.~prevalence-aware aggregation).  

\textbf{RQ2:} 
What happens when we change $g$ to equally or more valid measurement functions? 
In particular, we are interested in \emph{sensitivity}: for two measurements, $g_1$ and $g_2$, does $U_1$ correlate well with $U_2$; 
and \emph{external predictive validity}: for each measurement, $g_i$, does $U_i$ correlate well with the VIX, a stock-market volatility index based on S\&P 500 options prices? 

\citeauthor{baker2016measuring} also use the VIX as a measure of external validity, and like \citeauthor{baker2016measuring} we note that the VIX is a good proxy for \emph{economic uncertainty}, but does not necessarily capture \emph{policy} uncertainty. As \citeauthor{baker2016measuring} mention, ``differences in the topical scope between the VIX and the EPU index are an important source of distinct variation in the two measures." 
In the future, we could compare our measures to the other two external validity measures of \citeauthor{baker2016measuring}: mentions of \emph{uncertain} in the Federal Reserve's Beige Books and large daily moves in the S\&P stock index.




\textbf{Data and pre-processing.}  Although \citeauthor{baker2016measuring} use 10 newspapers to construct their US-based index, we instead use the New York Times Annotated Corpus (\emph{NYT-AC}) \cite{sandhaus2008new} because the text data is cleaned, easily accessible, and results on the corpus are reproducible. This collection includes over 1.8 million articles written and published by the New York Times between January 1, 1987 and June 19, 2007. \citeauthor{baker2016measuring} assume that using newspapers based in the United States is sufficient to find a signal of US-based EPU. To test this assumption, we apply a simple heuristic to the dateline of \emph{NYT-AC} articles and remove articles that mention non-US cities. However, we find relatively little variation in results via this heuristic (see Appendix, Figure~\ref{f:appendix-big-corr}).

\vspace{-0.2cm}
\subsection{Keyword matching}
Matching keyword lists, also known as \emph{lexicons} or \emph{dictionaries}, is a straightforward method to retrieve and/or classify documents of interest, and has the advantage of being interpretable. However, relying on a small set of keywords can create issues with recall and generalization.
On \emph{NYT-AC}, we apply the original keyword matching method of \citet{baker2016measuring} who label a document as positive if it matches any of 2 \emph{economy} keywords, AND any of 2 \emph{uncertainty} keywords, AND any of 13 \emph{policy} keywords, (\textbf{KeyOrg}). We also compare a method with the same economy and uncertainty matching criteria without \emph{policy} keyword matching (\textbf{KeyEU}); and a method for which we expand the economic and uncertainty keywords via word embeddings (\textbf{KeyExp}). 
See Table~\ref{t:epu-keywords} in the Appendix for the full list of keywords.  

\textbf{KeyExp}. Although \citeauthor{baker2016measuring} use human auditors to find \emph{policy} keywords that minimize the false positive and false negative rates, they do not expand or optimize for \emph{economy} or \emph{uncertainty} keywords. Thus, we expand these keyword lists via GloVe word embeddings\footnote{We used the 200-dimensional, 6B token corpus from Wikipedia and Common Crawl \url{http://nlp.stanford.edu/data/glove.6B.zip}} \cite{pennington2014glove}, and find the five nearest neighbors via cosine distance.\footnote{
We manually remove clear obvious negative keywords: \emph{policy} from the economic keyword bank and \emph{prospects} and \emph{remain} from the uncertainty keyword banks.}
This is a simple keyword expansion technique. In future work, one could look to the literature on \emph{lexicon induction} to improve creating lexicons that represent the semantic concepts of interest \cite{taboada2011lexicon,pryzant2018deconfounded,hamilton2016inducing,rao2009semi}. Alternatively, one could also create a probabilistic classifier over pre-selected lexicons to soften the predictions, or use other uncertainty lexicons or even automatic uncertainty cue detectors.  

\subsection{Document classifiers}
Probabilistic supervised machine learning classifiers are optimized to minimize the training loss between the predicted and true classes, and typically have better precision and recall trade-offs compared to keyword matching methods. We use 1844 documents and labels from \emph{BBD} from 1985-2007 as training data and 687 documents from 2007-2012 as a held-out test set. 
 We train a simple logistic regression classifier using \texttt{sklearn}\footnote{Version 0.22.1 \url{https://scikit-learn.org/}} \cite{pedregosa2011scikit} with a bag-of-words representation of text (\textbf{LogReg-BOW}). We tokenize 
and prune the vocabulary to retain words that appear in at least 5 documents, resulting in a vocabulary size of 15,968. We tune the L2-penalty via five-fold cross-validation. We also try alternative (non-BOW) text representations but these did not result in improved performance (Appendix, \S~\ref{s:appendix-measure}). Note that the labeled documents in \emph{BBD} are a biased sample as the authors select documents to annotate that match the \emph{economy} and \emph{uncertainty} keyword banks and do not select documents at random.

\vspace{-0.3cm} 
\subsection{Prevalence estimation}
Measuring economic policy uncertainty is an instance of  \emph{prevalence estimation}, the task of estimating the proportion of items in each given class.
Previous work has shown that simple aggregation methods over individual class labels can be biased if there is a shift in the distribution from training to testing or if the task is difficult \cite{keith2018uncertainty}. We compare aggregating via \textbf{classify and count (CC)}, taking the mean over binary labels, and \textbf{probabilistic classify and count (PCC)}, taking the mean over classifiers' inferred probabilities. See the Appendix \S\ref{ss:appendix-prev} for additional prevalence estimation experiments.  

\vspace{-0.2cm} 
\subsection{Results}
Addressing RQ2, our experimental results show that changes in measurement can result in substantial differences in the corresponding index. 
Table~\ref{t:doc-level} presents individual classification results on the training and test sets of \emph{BBD}, and Figures~\ref{f:time-series} and \ref{f:corr} show inference of the models on \emph{NYT-AC}. In Figure~\ref{f:time-series}, we note that the overall prevalences are substantially different: KeyExp has higher prevalence than KeyOrg as expected with more keywords but the supervised methods infer prevalences near 0.2 (CC) and 0.4 (PCC) which indicates they may be biased towards the training prevalence. 
 LogReg-BOW achieves both better individual classification predictive performance and combined with a probabilistic classify and count (PCC) prevalence estimation method achieves better correlation with the VIX (0.26 vs.~KeyOrg's 0.15). 
 The better predictive performance and correlation with VIX suggests PCC-LogReg-BOW represents a reasonable measurement of \emph{economic policy uncertainty}. Given this, the low correlation between PCC-LogReg-BOW and KeyOrg (0.38) raises concerning questions about KeyOrg's validity. 
 
 \vspace{-0.2cm}
 \subsection{Limitations} 
 {\tiny
\begin{table}[t]
\small
    \centering
    \begin{tabular}{l r}
    \toprule
     & Pearson's r \\
    \toprule
    KeyOrg-10 vs.~KeyOrg-NYT & 0.68 \\
    KeyOrg-10 vs.~VIX & 0.57 \\
    KeyOrg-NYT vs.~VIX & 0.15 \\
    \bottomrule
    \end{tabular}
    \caption{We use the official EPU index from \citeauthor{baker2016measuring} which applies keyword-matching (KeyOrg) on newspapers from 10 major outlets (10). For the years 1990-2006, we correlate this index with the same keyword-matching method on only the New York Times Annotated Corpus (NYT) and with the VIX.\label{t:off-corr}}
\end{table}
}

We use the \emph{NYT-AC} as a ``sandbox" for our experiments because of proprietary restrictions that limit us from acquiring the full text of all 10 news outlets used by \citeauthor{baker2016measuring} To understand the limitations of using only a single news outlet, we compare the ``official" aggregated index of \citeauthor{baker2016measuring}\footnote{From ``News\_Based\_Policy\_Uncert\_Index" column of \url{http://policyuncertainty.com/media/US_Policy_Uncertainty_Data.xlsx}} with KeyOrg applied to only the \emph{NYT-AC}. Table~\ref{t:off-corr} shows a 0.68 correlation between the official EPU index (KeyOrg-10) and the same keyword-matching method on only the NYT-AC (KeyOrg-NYT). Yet, KeyOrg-10 has a much higher correlation with the VIX, 0.57, compared to KeyOrg-NYT's correlation of 0.15. See Figure~\ref{f:epu-official} in the Appendix for a graph of these different indexes. We hypothesize applying PCC-LogReg-BOW to the texts of the all 10 newspapers used by \citeauthor{baker2016measuring} would result in improved external predictive validity, but we leave an empirical confirmation of this to future work. In practice, while keyword look-ups have lower recall than supervised methods they have the advantage of being interpretable and can use counts from document retrieval systems instead of full texts. 




\section{Related work}
There have been only a few other attempts to construct alternative, non-keyword measurements of \emph{economic policy uncertainty.}
\citet{azqueta2017developing} apply topic models and manually map the topics to \citeauthor{baker2016measuring}'s EPU categories and find their method tightly correlates (0.94) with the original index. 
In an unpublished manuscript, \citet{nyman2020text} expand the \emph{uncertainty} keywords of \citeauthor{baker2016measuring} via nearest neighbor embeddings and find Granger causality between their expanded keyword list and the original EPU index. In contrast, we are the first to take a fully supervised learning approach to measuring EPU and analyze the original annotations of \citeauthor{baker2016measuring}.

\textbf{Measurement of economic variables from text.} Other work has examined measuring economic variables from text data (see \citet{gentzkow2019text} for a survey). For example, topic models have been applied to central bank communications~\cite{hansen2018transparency} and newspaper articles~\cite{thorsrud2020words,bybee2020structure} while other work identifies negated uncertainty markers (e.g.~``there is \emph{no} uncertainty") in the Federal Reserve's Beige Books~\cite{saltzman2018machine} and extracts sentiment from central bank communications \cite{apel2012information}. \citet{boudoukh2019information} use off-the-shelf supervised document classifiers to demonstrate that the information in news can predict stock prices.







\textbf{Text-as-data methods.}
Traditional ways of analyzing textual data include \emph{content analysis} where human annotators read and hand-code documents for particular phenomena \cite{krippendorff2018content}. In the last decade, many researchers have adapted machine learning and NLP methods to the needs of social scientists \cite{card2019accelerating,o2011computational}. NLP technologies such as lexicons, topic models \cite{roberts2014structural, blei2003latent}, supervised classifiers, word embeddings \cite{mikolov2013distributed, pennington2014glove}, and large-scale pre-trained language model representations \cite{devlin2019bert} have been applied to textual data to extract relevant signals. More recent work attempts to extend text-as-data methods to incorporate principles from causal inference \cite{pryzant2018deconfounded,wood2018challenges,veitch2020adapting,roberts2020adjusting,keith2020text}.

\section{Future directions}
In the future, estimating the sensitivity of causal estimates to the different measurement approaches presented in this paper could potentially have substantive impact. Using a Bayesian modeling approach to annotator uncertainty \cite{paun2018comparing}, investigating better calibration, which has been shown to improve prevalence estimation \cite{card2018importance}, or estimating \emph{model uncertainty} could improve measurement. One could also shift from document-level predictions of EPU to paragraph, sentence, or span-level predictions. Annotating discourse structure and selecting discourse fragments, e.g.~\citet{Prasad04discourse}, could potentially increase annotator agreement. These sub-document extraction models could also potentially provide human-interpretable contextualization of movements in an EPU index.

\vspace{-0.2cm}
\section{Conclusion}
There is great promise for text-as-data methods and applications; however, we echo the cautionary advice of \citet{grimmer2013text} that automatic methods require extensive ``problem-specific validation.'' Our paper's investigation of \citeauthor{baker2016measuring} provides a number of general insights for text-as-data practioners along these lines.  
First, \emph{content validity:} when dealing with text data, one needs to think carefully about the kinds of linguistic information
one is trying to measure. For instance, mapping \emph{economic policy uncertainty} to a document-level binary label collapses all types of semantic uncertainty, many of which cannot be identified via keywords alone.
Second, one needs to examine \emph{social perception} assumptions. Is one trying to \emph{prescribe} an annotation schema, or, as we argue in this paper, are people’s perceptions about the concept as important as the concept itself, especially in the face of ambiguity in language? 
Third, \emph{sensitivity of measurements:} text-as-data practitioners can strengthen their substantive conclusions if multiple measurement approaches give similar results. For \emph{economic policy uncertainty}, this paper demonstrates that switching from keywords to aggregating the outputs of a document classifier are not tightly correlated, a concerning implication for the validity of this index. 

\section*{Acknowledgements}
The authors thank Bloomberg's AI Engineering team, especially Diego Ceccarelli, Miles Osborne and Anju Kambadur, as well as Su Lin Blodgett for helpful feedback and directions. Additional thanks to the anonymous reviewers from the 2020 Natural Language Processing and Computational Social Science Workshop for their insights. Katherine Keith acknowledges support from Bloomberg's Data Science Ph.D. Fellowship.

\bibliographystyle{acl_natbib}
\bibliography{emnlp2020}

\appendix
\section*{Appendix}

\section{Datasets} \label{s:datasets}

Here we provide more information on the data used in annotation and measurement experiments. 

\begin{itemize}
    \item \emph{BBD.} From \  \citet{baker2016measuring}, we combine the authors' annotations with the full text data they provided.\footnote{\url{http://policyuncertainty.com/AUDIT_ANALYSIS.zip} and \url{http://policyuncertainty.com/media/All\%20Audit\%20Hard\%20Copies.rar}} These documents and annotations are sampled from ten major newspapers in the United States.\footnote{LA Times, USA Today, Chicago Tribune, Washington Post, Boston Globe, Wall Street Journal, New York Times, Miami Herald, Dallas Morning News, San Francisco Chronicle} We also study and refer to their \emph{Code Guide} when analyzing examples for this paper.\footnote{\url{http://policyuncertainty.com/media/Coding_Guide.pdf}} See Lines 1-2 of Table~\ref{t:ann-exp} for descriptive statistics of this dataset.  
    
    \item \emph{NYT-AC.} We use the \emph{New York Times Annotated Corpus} as a sandbox for our experiments \cite{sandhaus2008new}.\footnote{\url{https://catalog.ldc.upenn.edu/LDC2008T19}} 
    This corpus includes over 1.8 million articles written and published by the New York Times between January 1, 1987 and June 19, 2007.
    \item \emph{VIX.} The VIX is an index of market expectations of the next 30 days' U.S.\ stock market volatility,
derived from S\&P 500 options prices. Like \citeauthor{baker2016measuring}, we take a monthly average over the daily VIX measures, obtained from a standard proprietary database.  
\end{itemize}

\section{Annotation notes}
\begin{figure}[h!]
\centering
\includegraphics[width=0.95 \columnwidth]{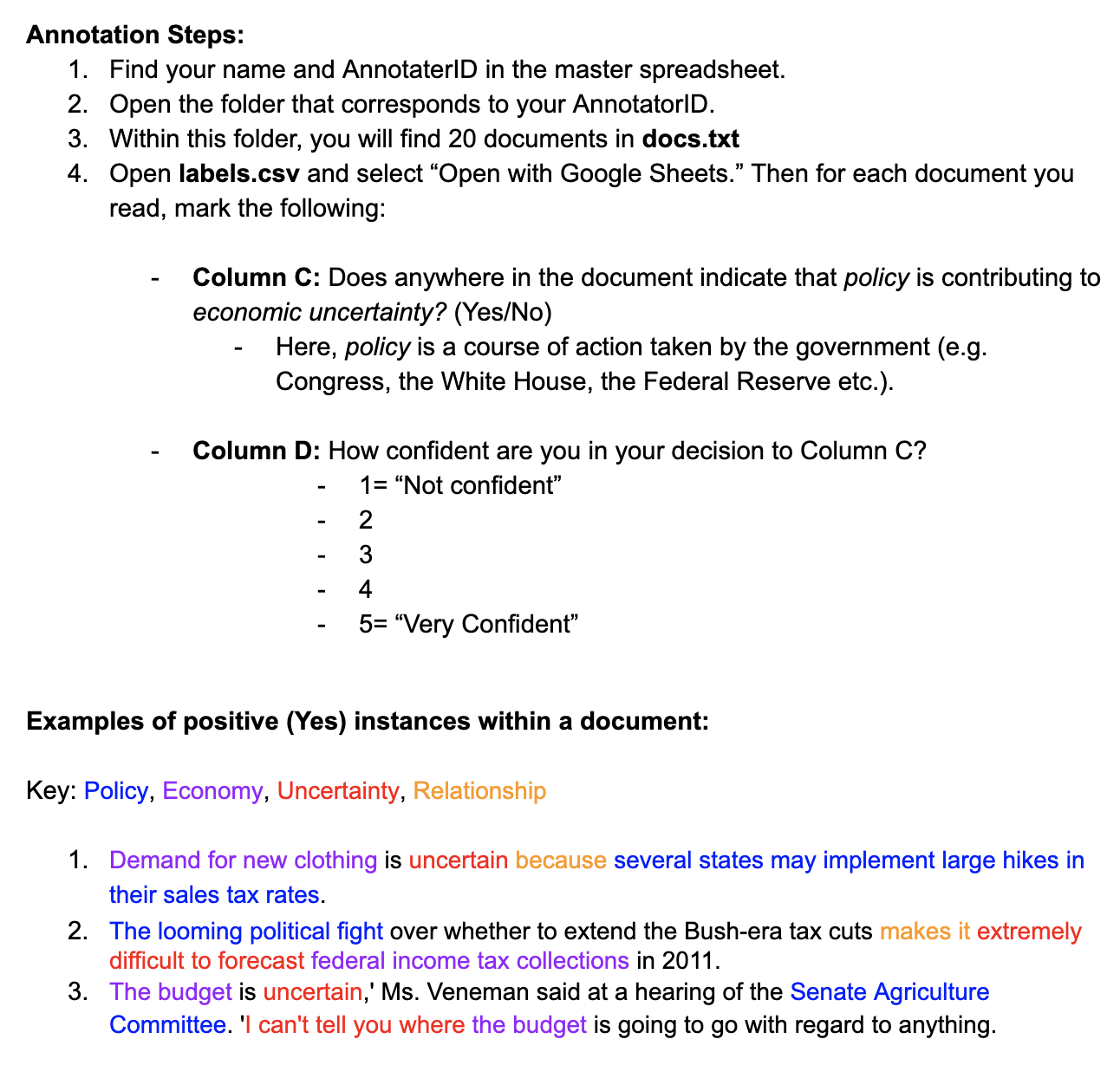}
\caption{Annotation instructions for our experiment. \label{f:ann-instruct}}
\end{figure}

We provide additional descriptive statistics of \citet{baker2016measuring}'s original annotations in Tables~\ref{t:per-ann-stats} and \ref{t:num-ann-per-doc}. The annotation instructions for our experiment (\S\ref{ss:our-anns}) are provided in Figure~\ref{f:ann-instruct}. In our annotation experiment, the mean annotator confidence levels are 3.81 for Sample A and 3.85 for Sample B. 

\begin{table}[h!]
  \centering
      \begin{tabular}{rrrr}
\toprule Annotator & Mean pos. & Std & N \\ \toprule
A & 0.300 & 0.46 & 50 \\
B & 0.619 & 0.49 & 278 \\
C & 0.431 & 0.50 & 297 \\
D & 0.445 & 0.50 & 449 \\
E & 0.551 & 0.50 & 472 \\
F & 0.330 & 0.47 & 790 \\
G & 0.402 & 0.49 & 1168 \\
H & 0.389 & 0.49 & 1185 \\
I & 0.350 & 0.48 & 1265 \\
J & 0.398 & 0.49 & 1443 \\
K & 0.455 & 0.50 & 1606 \\
L & 0.397 & 0.49 & 1740 \\
M & 0.405 & 0.49 & 2320 \\
N & 0.443 & 0.50 & 2997 \\
  \end{tabular}
  \caption{Descriptive statistics for the original annotations of \citet{baker2016measuring}. Annotator names have been anonymized to letters.  For each annotator, we report the mean number of positive annotations (mean pos.), the standard deviation of positive annotations (std), and the total number of annotations by that annotator (N).  \label{t:per-ann-stats}}
\end{table}

\begin{table}[h!]
    \centering
    \begin{tabular}{rr}
\toprule Num. Annotators & Num. Docs \\ \toprule
1 & 11647 \\
2 & 2053 \\
3 & 83 \\
4 & 12 \\
5 & 2 \\
    \end{tabular}
    \caption{For \citeauthor{baker2016measuring}'s original dataset, the number of documents that have a particular number of annotators. Here, 16\% of documents have only a single annotator. 
    \label{t:num-ann-per-doc}}
\end{table}

\section{Qualitative examples}
Figures~\ref{t::agreement_examples} and \ref{t::disagreement_examples} provide examples with high annotator agreement and disgreement respectively.  

\begin{table*}[h!]
\small
  \centering
      \begin{tabularx}{\linewidth}
      {lXX
      l}
      \toprule
      &Example selection & Our analysis & Label Mean, Docid \\
      \toprule
      1 & \dots Several recent news reports have questioned the stamina of Wells Fargo's real estate portfolio in the event of a recession that extends to California. The analysis had driven the bank's stock sharply down. \dots. 
      & Stock market newsletter digest. Economics policy is not mentioned as uncertain
      & 0.0,~MIHB\_11\_1990\_8 \\ 
      \midrule
      2 & \dots Just eight days before the threatened imposition of punitive U.S. tariffs on Japanese luxury cars, Japanese automakers are signaling a strong desire to compromise with Washington in the bitter dispute over automotive trade. \dots & Report on international trade dispute. "threatened" directly expresses uncertainty, "tariffs" are economic policy & 0.8,~DMNB\_6\_1995\_8 \\
      \bottomrule
  \end{tabularx}
  \caption{
        Hand-selected examples with strong annotator agreement. Docids correspond to those provided in  \citet{baker2016measuring}'s dataset. Label mean is the mean over our experiment's five annotations per document.  
        \label{t::agreement_examples}
    }
\end{table*}

\begin{table*}[h!]
\small
  \centering
      \begin{tabularx}{\linewidth}{lXXl}
      \toprule
      &Example selection & Our analysis & Label Mean, Docid \\
      \toprule
      1 &\dots I am a true believer that mobile broadband will help my company and hundreds of other businesses in South Florida work more efficiently, better serve consumers and hire more employees. On a related matter, policymakers in Washington, D.C. are making decisions on whether to allow AT\&T to pay approximately \$39 billion for its wireless rival T-Mobile. This is a deal of vital importance to our community \dots. 
      & An op-ed arguing that a merger should be allowed to go forward. Arguing for a certain outcome implies uncertainty of the outcome, but uncertainty is never explicitly stated.
      & 0.4,~MIHB\_5\_2011\_3 \\ 
      \midrule
      2 &\dots angst over rising interest rates triggered a nasty sell-off in the stock market Friday \dots The markets also fret that the Federal Reserve Board will move to curb that inflation threat \dots 
      & Reports on downturn in stock market. Annotators must decide: is there uncertainty about FED actions or strong expectation of disfavoured actions. & 0.4,~LA\_8\_1997\_9 \\
      \midrule
      3 & \dots If Cuba's fledgling recovery is to continue, Mr. Castro must legalize small- and medium-sized businesses, boost wages and gradually introduce free markets, U.S. officials say. \dots Cuban officials have a very different view and blame the long-time U.S. ban on trade with the island for much of their economic woes.\dots & Reports on state of affairs in Cuba. States assumption that US or Cuban policy will eventually lead to economic problems. Uncertainty is only implied and no concrete policies are mentioned  & 0.6,~DMNB\_12\_1999\_2 \\
      \midrule
      4 & \dots Two military coups and several attempts, race riots and poverty have made the Kingdom in the Sky a place of turmoil in the past years. Economic problems and the repeal of apartheid in South Africa, Lesotho's overpowering neighbor on all sides, raise even more questions about what lies ahead. Sympathetic foreign powers have donated millions to Lesotho. \dots 
      & Describes situation in Lesotho. Mentions economic downturn, large turnover in administrations and race riots. Not stated that turnover/riots lead to uncertainty over economic policy, but could be reasonably inferred as part reason for downturn. & 
      0.4,~MIHB\_7\_1991\_15 \\
      \bottomrule
  \end{tabularx}
   \caption{
        Hand-selected examples with strong annotator disagreement. Docids correspond to those provided in  \citet{baker2016measuring}'s dataset.  Label mean is the mean over our experiment's five annotations per document. 
        \label{t::disagreement_examples}
    }
\end{table*}

\section{Measurement: Additional Experiments} \label{s:appendix-measure} 

In this section, we provide additional measurement experiments. Also note there is a very small overlap between our training time documents and inference time NYT-AC documents. There are 375 documents at training time from NYT between the years of 1990 and 2006. However, the total number of inference documents is 1,501,131 so this is less than 0.025\% of documents.

\subsection{Filtering to US-Only News}
Initial qualitative analysis reveals that many documents, and in particular articles with high annotator disagreement, are focused on events outside the United States. An unstated assumption of \citet{baker2016measuring} is that US-based news sources will primarily report US-based news and thus US-based economic policy uncertainty. We test this assumption empirically. 

To remove non-US news, we use a simple heuristic that gives almost perfect precision.
\emph{NYT-AC} has metadata about the dateline of an article, for example ``KUWAIT, Sunday, March 30," ``SAN ANTONIO, March 29," or ``BAGHDAD, Iraq, March 29." We (1) use the \emph{GeoNames} Gazateer\footnote{\url{http://www.geonames.org/}} and filter to cities that have greater than 15,000 inhabitants;\footnote{\url{https://datahub.io/core/world-cities}} (2) separate these city names into US and non-US cities such that ties go to US. 
For example, \emph{Athens} would not be removed because the town of Athens, Georgia is in the United State; (3) write a rule-based text parser that extracts the span of text that is in all capitals, (4) if the city name is in non-US cities, we discard the document. 

\begin{figure*}[t]
\centering
\includegraphics[width=0.9\textwidth]{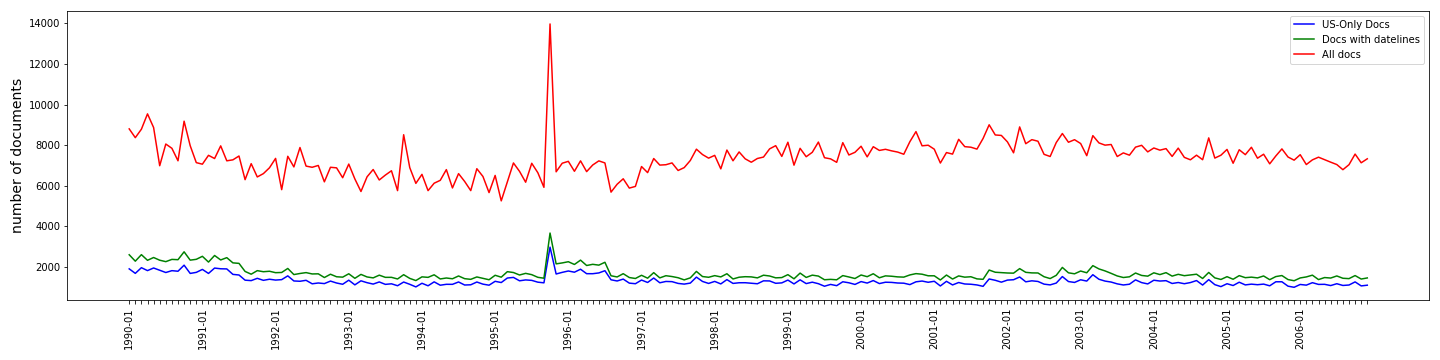}
\caption{ NYT total documents (red), documents with datelines (green) and documents for which the dateline does not have a non-US city (blue). We checked and confirmed and the spike in 1995-10 is an artifact of the corpus. \label{f:nyt-all-docs}}
\end{figure*}

Per month, on average, we remove 449 documents that were about non-US news. 
See Figure~\ref{f:nyt-all-docs} for a comparison of all NYT articles, articles with the dateline, and US-only articles based on our heuristic. 

Figure~\ref{f:appendix-big-corr} displays correlation results for all models with the US-Only document filter. Applying the US-Only filter only slightly improves correlation of all models with the VIX (0.01-0.04 correlation). From these results, it seems that \citeauthor{baker2016measuring}'s assumption is valid. However, we also acknowledge that our heuristic is high-precision, low recall and in the future, one could possibly use a country-level document classifier instead. 


\begin{figure}[t]
\centering
\includegraphics[width=0.75\columnwidth]{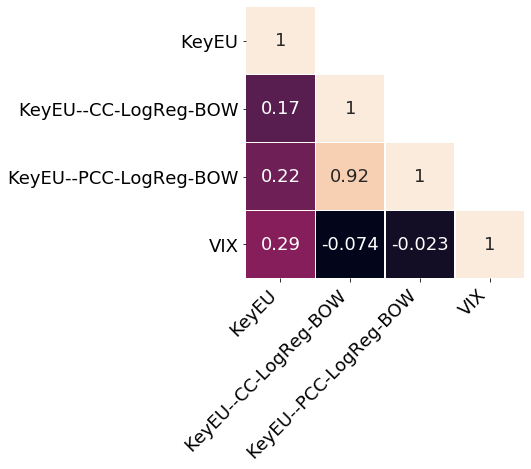}
\caption{ Estimate PCC and CC only within the set of documents that pass the EU filter. \label{f:eu-only-doc-class}}
\end{figure}


\subsection{Predicting after EU filter}
As we acknowledge in the main text, the training set is biased because documents were sampled only if they matched the \emph{economy} and \emph{uncertainty} keyword banks. To make a fair comparison at inference time, we looked at the predictions of our document classifiers on the subset of documents in \emph{NYT-AC} that also matched these \emph{economy} and \emph{uncertainty} keyword banks (KeyEU). In Figure~\ref{f:eu-only-doc-class}, we see that the subset of these models had lower correlation with the VIX.  

\begin{table*}[t]
  \centering
      \begin{tabular}{llrrrrr}
      \toprule
      Split &Model & Prec. & Recall & F1 & Acc. \\
      \toprule
Train & LogReg-BERT&0.79&0.77&0.78&0.79 \\ 
\hline  
Test & LogReg-BERT&0.61&0.59&0.60&0.68 \\ 
      \bottomrule
  \end{tabular}
  \caption{Performance results on the training and test sets for the \emph{LongFormer} representation with logistic regression (LogReg-BERT). The results in this table are comparable to Table~\ref{t:doc-level}. \label{t:doc-level-appdx}}
\end{table*}

\subsection{Additional prevalence estimation experiments}\label{ss:appendix-prev}

\begin{figure*}[h]
\centering
\includegraphics[width=0.8\textwidth]{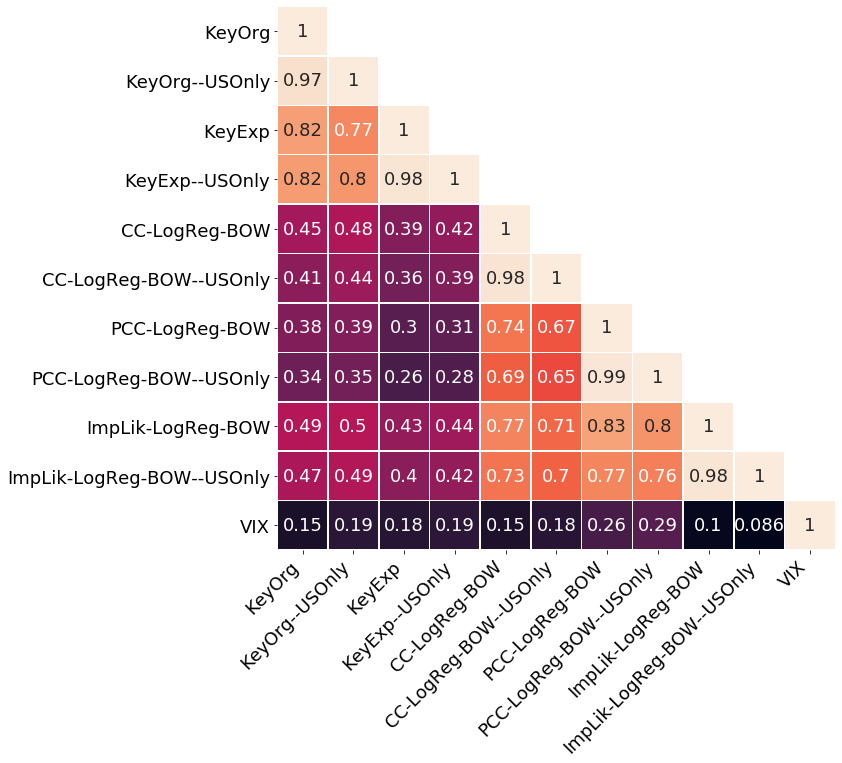}
\caption{Correlations between all models. The addition of \emph{--USOnly} to a model name means we apply the model only on the subset of documents that have passed our USOnly heuristic. \emph{ImpLik} is the implicit likeihood prevalence estimation method of \citet{keith2018uncertainty}. \label{f:appendix-big-corr}}
\end{figure*}

As an alternative to classify and count (CC) and  probabilistic classify and count (PCC) prevalence estimation methods, we also experiment with the \emph{Implicit Likelihood (ImpLik)} prevalence estimation method of \citet{keith2018uncertainty}. This method gives the predictions of a discriminative classifier a generative re-interpretation and backs out an implicit individual-level likelihood function which can take into account bias in the training prevalence. We use the authors' \texttt{freq-e} software package.\footnote{\url{https://github.com/slanglab/freq-e}. For the label prior we used the training prevalence of 0.48. 
} Figure~\ref{f:appendix-big-corr} shows a high correlation between \emph{ImpLik} and \emph{PCC}, 0.83 correlation; however, \emph{ImpLik} had much lower correlation with the VIX (0.1). Note, the mean prevalences from \emph{ImpLik} are much lower than PCC or CC with a mean monthly prevalence across 1990-2006 of 0.02. 
Thus, the method seems to be correcting for a more realistic prevalence but the true prevalence values may be too low to pick-up relevant signal via this method. 

\vspace{-.3cm}
\subsection{BERT representations}
Finally, we acknowledge that a bag-of-words representation in the document classifier is dissatisfying to capture long-range semantic dependencies and the contextual nature of language that has motivated recent research in contextual, distributed representations of text. Thus, we use the frozen representations of a large, pre-trained language model that has been optimized for long documents, the  \emph{LongFormer} \cite{beltagy2020longformer}. This is a model that optimizes a RoBERTa model \cite{liu2019roberta} for long documents. 
We use the \texttt{huggingface} implementation of the LongFormer\footnote{\url{https://huggingface.co/transformers/model_doc/longformer.html}} and use the 768-dimensional ``pooled output"\footnote{This is the hidden state of the last layer of the first token of the sequence which is then passed through a linear layer and Tanh activation function. The linear layer weights are trained from the next sentence prediction objective during pre-training.} as our document representation. We then use the same \texttt{sklearn} logistic regression training as the BOW models. 

Comparing Table~\ref{t:doc-level-appdx} to Table~\ref{t:doc-level}, we see that this representation has decreased performance compared to LogReg-BOW. We speculate that this decrease in performance may originate in having to truncate documents to 4096 tokens due to the constraints of the model architecture. With more computational resources, we would fine-tune the pre-trained weights instead of leaving them frozen. Future work could also consider obtaining alternative representations of text via weighted averaging of embeddings \cite{arora2017simple}, deep averaging networks \cite{iyyer2015deep}, or pooling BERT embeddings of all paragraphs in a document.  

\begin{figure*}[h]
\centering
\includegraphics[width=0.9\textwidth]{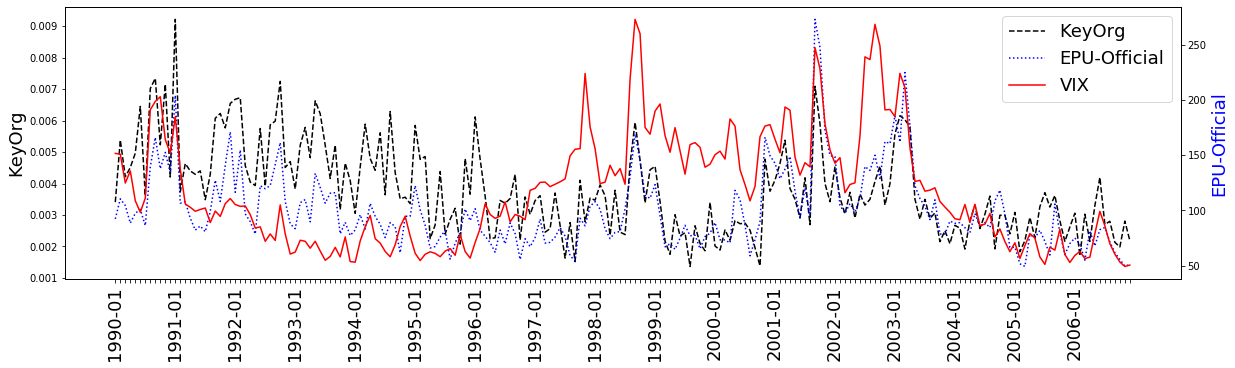}
\caption{Official EPU versus the original keywords on the \emph{NYT-AC} (KeyOrg). \label{f:epu-official}}
\end{figure*}

\end{document}